\documentclass[11pt,a4paper]{article}
\usepackage[hyperref]{acl2021}
\usepackage{threeparttable, subfigure}
\usepackage{graphicx}
\usepackage{times}
\usepackage{latexsym}

\usepackage{multirow}
\usepackage{multicol}
\usepackage{booktabs}
\usepackage{adjustbox}
\usepackage{enumitem}
\usepackage{bm}
\usepackage{amsmath,amsfonts,amssymb}
\newcommand\mymathop[1]{\mathop{\operatorname{#1}}}
\DeclareMathOperator*{\argmax}{arg\,max}
\usepackage{bbm}
\usepackage{tikz}

\usepackage{pifont}
\usepackage{microtype}
\usepackage{cleveref}
\crefname{section}{§}{§§}
\Crefname{section}{§}{§§}
\aclfinalcopy 

\title{Fully Non-autoregressive Neural Machine Translation: 
\\Tricks of the Trade}
\author{Jiatao Gu\thanks{\quad Equal contribution.}\\
  Facebook AI Research \\
  \texttt{jgu@fb.com} \\\And
  Xiang Kong\footnotemark[1]\\
  Language Technologies Institute \\
  Carnegie Mellon University\\
  \texttt{xiangk@cs.cmu.edu} \\}

\date{}

\begin{document}
\maketitle
\begin{abstract}
Fully non-autoregressive neural machine translation (NAT) is proposed to
simultaneously predict tokens with single forward of neural networks, which significantly reduces the inference latency at the expense of quality drop compared to the Transformer baseline. 
In this work, we target on closing the performance gap while maintaining the latency advantage. 
We first inspect the fundamental issues of fully NAT models, and adopt \textit{dependency reduction} in the learning space of output tokens as the basic guidance. Then, we revisit methods in four different aspects that have been proven effective for improving NAT models, and carefully combine these techniques with necessary modifications. Our extensive experiments on three translation benchmarks show that the proposed system achieves the new state-of-the-art results for fully NAT models, and obtains comparable performance with the autoregressive and iterative NAT systems. For instance, one of the proposed models achieves $\bm{27.49}$ BLEU points on WMT14 En-De with approximately $\bm{16.5\times}$ speed up at inference time.

\end{abstract}

\section{Introduction} \label{introduction}
\begin{figure}[t]
    \centering
    \includegraphics[width=\linewidth]{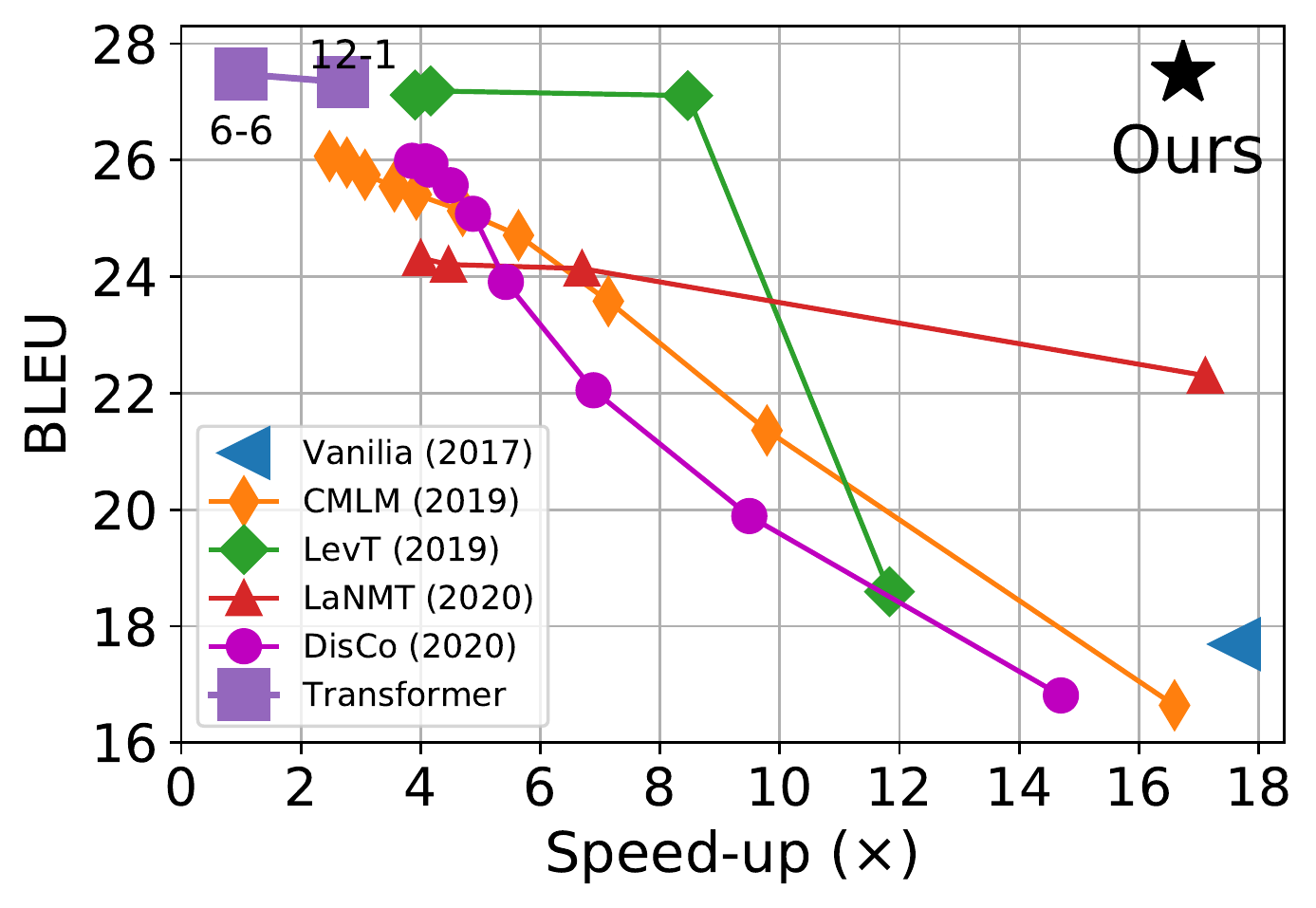}
    \caption{The translation quality v.s. inference speed-up of the proposed model with the AT (Transformer) and existing popular iterative NAT models varying decoding iterations on WMT14 En-De test set. The upper right corner achieves the best trade-off.\vspace{-10pt}} 
    \label{fig:example}
\end{figure}
State-of-the-art neural machine translation (NMT) systems are based on autoregressive models~\cite{bahdanau2014neural,vaswani2017attention} where each generation step depends on the previously generated tokens. Such sequential nature inevitably leads to inherent latency at inference time.
Non-autoregressive neural machine translation models~\cite[NAT,][]{gu2017non} attempt to generate output sequences in parallel to speed-up the decoding process. The incorrect independence assumption nevertheless prevents NAT models to properly learn the dependency between target tokens in real data distribution, resulting in a performance drop compared to autoregressive (AT) models. One popular solution to improve the translation accuracy is to sacrifice the speed-up by incorporating an iterative refinement process in NAT models, through which the model explicitly learns the conditional distribution over partially observed reference tokens. However, recent studies~\cite{kasai2020deep} indicated that 
iterative NAT models seem to lose the speed advantage after carefully tuning the layer allocation of AT models. For instance, an AT model with \textit{deep encoder and shallow decoder} (12-1) obtains similar latency as iterative NAT models without hurting the translation accuracy.

Several works~\cite{GhazvininejadKZ20,saharia-etal-2020-non,qian2020glancing} have recently been proposed to improve the fully NAT models, though the performance gap compared to the iterative ones remains. How to build a fully NAT model with competitive translation accuracy calls for more exploration. In this work, we first argue that the key to successfully training a fully NAT model is to perform \textit{dependency reduction} in the learning space of output tokens (\cref{background}). With this guiding principle, we revisit various methods which are able to reduce the dependencies among target tokens as much as possible from four different perspectives, i.e., training corpus (\cref{3_1}), model architecture (\cref{3_2}), training objective (\cref{3_3}) and learning strategy (\cref{3_4}). Furthermore, such target token dependencies cannot be perfectly removed by any of these aspects only. The performance gap can not be near closed unless we make full use of these techniques' advantages. 

We validate our fully NAT model on standard translation benchmarks including $5$ translation directions. The proposed system achieves new state-of-the-art results for fully NAT models. Moreover, compared to the Transformer baseline, our models achieve comparable performance with over \textbf{16$\times$} speed-up at inference time.

\section{Motivation} \label{background}

Given an input sequence $\bm{x}=x_1\ldots x_{T'}$, an autoregressive model~\cite{bahdanau2014neural,vaswani2017attention} predicts the target $\bm{y}=y_1\ldots y_{T}$ sequentially based on the conditional distribution $p(y_t|y_{<t}, x_{1:T'}; \theta)$, which tends to suffer from high latency in generation especially for long sequences. In contrast, non-autoregressive machine translation~\citep[NAT,][]{gu2017non}, proposed for speeding-up the inference by generating all the tokens in parallel, has recently been on trend due to its nature of parallelizable on devices such as GPUs and TPUs. A typical NAT system assumes a conditional independence in the output token space, that is
\begin{equation}
    p_\theta(\bm{y} | \bm{x}) = \prod_{t=1}^T p_\theta(y_t|x_{1:T'})
    \label{eq.nat}
\end{equation}
where $\theta$ is the parameters of the model. Typically, NAT models are also modeled with Transformer encoder-decoder without causal attention map in the decoder side.   
However, as noted in~\citet{gu2017non}, the independence assumption generally does not hold in real data distribution for sequence generation tasks such as machine translation~\cite{ren2020study}, where the failure of capturing such dependency between target tokens leads to a serious performance degradation in NAT. 

\begin{figure}
    \centering
    \includegraphics[width=\linewidth]{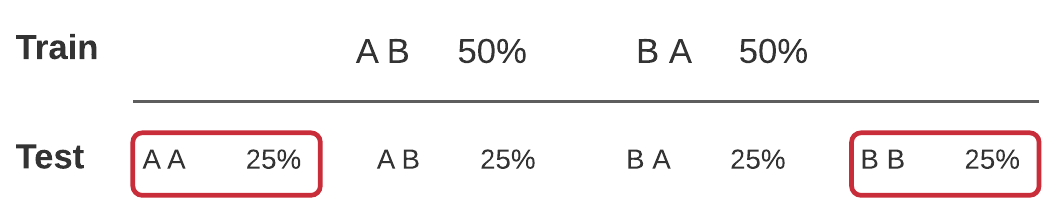}
    \caption{Toy example shows that NAT fails to learn when dependency exists in output space.}
    \label{fig:toy}
\end{figure}

This is a fairly understandable but fundamental issue of NAT modeling which can be easily shown with a toy example in Figure~\ref{fig:toy}. Given a simple corpus with only two examples: \textit{AB} and \textit{BA}, each of which has $50\%$ chances to appear. It is designed to represent the dependency that symbol \textit{A} and \textit{B} should co-occur. Although such simple distribution can be instantly captured by any autoregressive model, learning the vanilla NAT model with maximum likelihood estimation (MLE, Eq.~\eqref{eq.nat}) assigns probability mess to incorrect outputs (\textit{AA}, \textit{BB}) even these samples never appear during training.
In practice, the dependency in real translation corpus is much more complicated.
As shown in Figure~\ref{fig:example}, despite the inference speed-up, 
training the vanilla NAT the same as Transformer in a comparable size leads to quality drop over $\textbf{10}$ BLEU points.

To ease the modeling difficulty, recent state-of-the-art NAT systems~\cite{lee2018deterministic,stern19a,ghazvininejad2019mask,gu2019levenshtein,kasai2020non,shu2020latent,saharia-etal-2020-non} trade accuracy with latency by incorporating an iterative refinement process in non-autoregressive prediction, and have already achieved comparable or even better performance than the autoregressive counterpart. Nevertheless, \citet{kasai2020deep} showed autoregressive models with a deep encoder and a shallow decoder can readily outperform strong iterative models with similar latency, indicating that the latency advantage of iterative NAT has been overestimated.

By contrast, while maintaining a clear speed advantage, fully NAT system -- model makes parallel predictions with single neural network forward -- still lags behind in translation quality and has not been fully explored in literature~\cite{libovicky-helcl-2018-end,li2018hint,sun2019fast,ma2019flowseq,GhazvininejadKZ20}. This motivates us in this work to investigate various approaches to push the limits of learning a fully NAT model towards autoregressive models regardless of the architecture choices~\citep{kasai2020deep}.


\section{Methods} \label{method}
\begin{figure*}[t]
    \centering
    \includegraphics[width=1.0\linewidth]{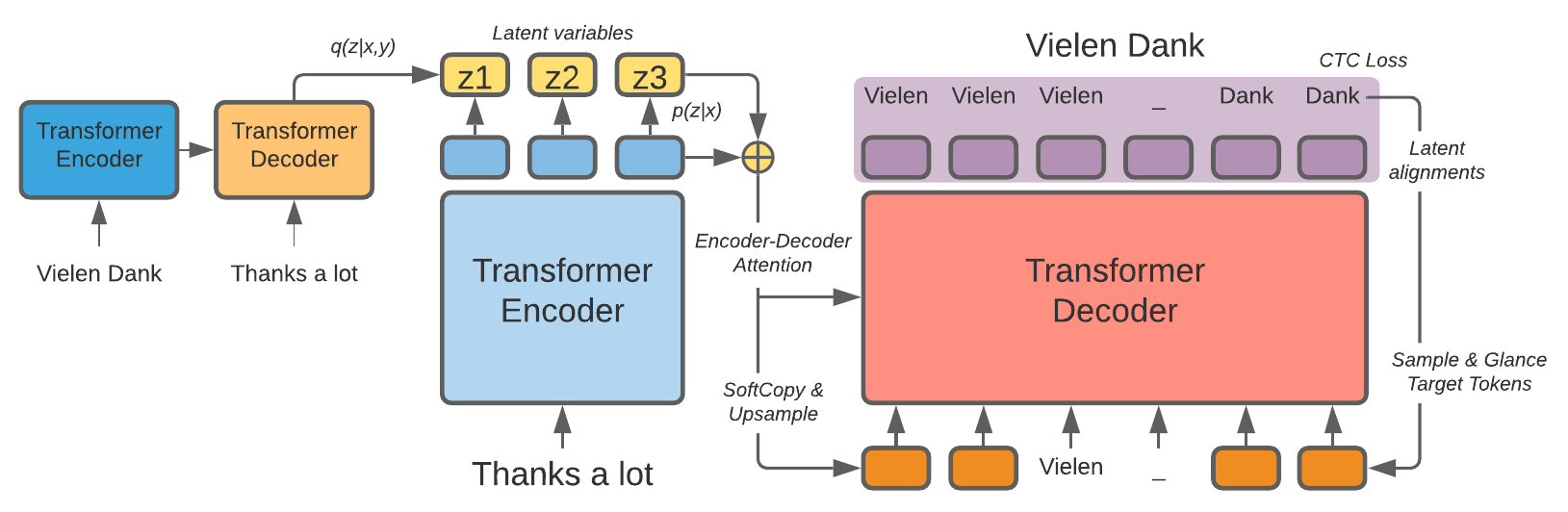}
    \caption{The overall framework of our fully NAT model.}
    \label{fig:framework}
\end{figure*}
In this section, we discuss several important ingredients to train a fully NAT model. As discussed in \cref{background}, we argue that the guiding principle of designing any NAT models is to perform \textit{dependency reduction} as much as possible in the output space so that it can be captured by the NAT model. For example, iterative-based models~\cite{ghazvininejad2019mask} explicitly reduce the dependencies between output tokens by learning the conditional distribution over the observed reference tokens. The overall framework of training our fully NAT system is presented in Figure~\ref{fig:framework}.

\begin{table*}[ht!]
  \caption{\label{Tab:bookRWCal}Comparison between the proposed techniques for improving fully NAT models}
  \centering
  \small
  \begin{tabular}{lp{2.9cm}p{2.9cm}p{2.9cm}p{2.9cm}}
  \toprule
  \textbf{Methods} &\textbf{Distillation} &\textbf{Latent Variables}&\textbf{Latent Alignments}&\textbf{Glancing Targets}\\
  \midrule
  What it can do?  & simplifying the training data & model any types of dependency in theory & handling token shifts in the output space & ease the difficulty of learning hard examples\\
  \midrule
  What it cannot?  & uncertainty exists in the teacher model & constrained by the modeling power of the used latent variables & unable to model non-monotonic dependency, e.g. reordering & training / testing phase mismatch\\
  \midrule
  Potential issues & sub-optimal due to the teacher's capacity & difficult to train; posterior collapse & decoder inputs must be longer than targets & difficult to find the optimal masking ratio\\
  \bottomrule
  \end{tabular}
\end{table*}
\subsection{Data: Knowledge Distillation} \label{3_1}
The most effective \textit{dependency reduction} technique is knowledge distillation (KD)~\cite{hinton2015distilling,kim2016sequence} which is firstly proposed in~\citet{gu2017non} and has been widely employed for all subsequent NAT systems. We replace the original target samples with sentences generated from a pre-trained autoregressive model. As analyzed in \newcite{zhou2019understanding}, KD is able to simplify the training data where the generated targets have less noise and are aligned to the inputs more deterministically. \newcite{zhou2019understanding} also showed that the capacity of the teacher model should be constrained to match the desired NAT model to avoid further degradation in KD, especially for weak NAT students without iterative refinement. We treat Transformer \textit{base} as the default teacher model to generate the distilled corpus.

\subsection{Model: Latent Variables}\label{3_2}
In contrast to iterative NAT, \textit{dependency reduction} can be done with (barely) zero additional cost at inference time by utilizing latent variables as part of the model. In such cases, output tokens $y_{1:T}$ are modeled conditionally independent over the latent variables $\bm{z}$ which are typically predicted from the inputs $x_{1:T'}$:
\begin{equation}
    p_\theta(\bm{y}|\bm{x})=\int_{\bm{z}}p_\theta(\bm{z}|\bm{x})\prod_{t=1}^Tp_\theta( y_t| \bm{z}, \bm{x})d\bm{z}
\end{equation}
where $z$ can be pre-defined by external library (e.g. fertility in~\citet{gu2017non}), or jointly optimized with the NAT model using normalizing flow~\citep{ma2019flowseq} or variational auto-encoders (VAEs)~\cite{kaiser2018fast,shu2020latent}. 

In this work, we followed the formulation proposed in \newcite{shu2020latent} where continuous latent variables $\bm{z} \in \mathbb{R}^{T'\times D}$ are modeled as spherical Gaussian at the encoder output of each position. Like standard VAEs~\cite{kingma2013auto}, we train such model to maximize the evidence lower-bound (ELBO) with a posterior network $q_\phi$:
\begin{equation}\label{eqn:vae}
    \underbrace{\mymathop{\mathbb{E}}_{\bm{z}\sim q_\phi}\left[\log p_\theta(\bm{y}|\bm{z}, \bm{x})\right]}_{\text{likelihood}} - \mathcal{D}_{\text{KL}}(q_\phi(\bm{z}|\bm{x}, \bm{y})\|p_\theta(\bm{z}|\bm{x}))
\end{equation}
where we use an encoder-decoder architecture similar to the main model to encode $q_\phi(\bm{z}|\bm{x}, \bm{y})$. In our implementation, only the parameters of the embedding layers are shared between $\theta$ and $\phi$

\subsection{Loss Function: Latent Alignments}\label{3_3}
Standard sequence generation models are trained with the cross entropy (CE) loss which compares model's output with target tokens at each corresponded position. However, as NAT ignores the dependency in the output space, it is almost impossible for such models to accurately model token offset. For instance, while with little effect to the meaning, simply changing \textit{Vielen Dank !} to \textit{, Vielen Dank} causes a huge penalty for fully NAT models.

To ease the limitations in loss computation, recent works have proposed to consider latent alignments between the target positions, and ``actively'' search the best~\citep[AXE,][]{GhazvininejadKZ20} or marginalize all the alignments~\citep[CTC,][]{libovicky-helcl-2018-end,saharia-etal-2020-non} with dynamic programming. The dependency is reduced because the NAT model is able to freely choose the best prediction regardless of the target offsets. 

In this work, we put our major focus on the
CTC loss~\cite{graves2006connectionist} considering its superior performance and the flexibility of variable length prediction. 
Formally, 
given the conditional independence assumption,
CTC is capable of efficiently finding all valid aligned sequences $\bm{a}$ which the target $\bm{y}$ can be recovered from, and marginalize log-likelihood:
\begin{equation}
    \log p_{\theta}(\bm{y}|\bm{x}) = \log \sum_{\bm{a} \in \Gamma(\bm{y})} p_{\theta}(\bm{a}|\bm{x})
\end{equation}
where $\Gamma^{-1}(\bm{a})$ is the collapse function that recovers the target sequence by
collapsing consecutive repeated tokens, and then removing all blank tokens. 
Furthermore, it is straightforward to apply the same CTC loss into latent variable models by replacing the likelihood term in Eq~\eqref{eqn:vae} with the CTC loss. Note that both CTC and AXE make strong  assumptions of monotonic alignment, which makes them impossible to reduce all dependencies between target tokens in real distribution.

\subsection{Learning: Glancing Targets}\label{3_4}
Although it is possible to directly optimize the fully NAT model towards the target sequence $\log p_\theta(\bm{y}|\bm{x})$ (or ELBO if using latent variables), \newcite{ghazvininejad2019mask} showed that it improved test time performance by training the NAT model with randomly sampled reference tokens as inputs $\log p_\theta(\bm{y}|\bm{m}\odot\bm{y}, \bm{x}), \bm{m} \sim \gamma(l, \bm{y}), l \sim \mathcal{U}_{| \bm{y} |}$, where $\bm{m}$ is the mask, and $\gamma$ is the sampling function given the number of masked tokens $l$.
As mentioned earlier, we suspect such explicit modeling of the
distribution conditional to observed tokens assists the \textit{dependency reduction} in the output space.
\paragraph{Curriculum Learning}
Naively applying random masks for every training example may cause severe mismatch between training and testing.
\newcite{qian2020glancing} proposed GLAT -- a curriculum learning 
strategy, in which the ratio of observed target tokens is related to the inference quality of the fully NAT model. More precisely, instead of sampling uniformly, we sample $l$ by:
\begin{equation}
    l \sim g(f_{\text{ratio}} \cdot \mathcal{D}(\bm{\hat{y}}, \bm{y}))
\end{equation}
where $\bm{\hat{y}}=\argmax_{\bm{y}}\log p_\theta(\bm{y}|\bm{x})$, $\mathcal{D}$ is the discrepancy  between the model prediction and the target sequence, e.g. 
Levenshtein distance~\cite{levenshtein1966binary}, and $f_{\text{ratio}}$ is a hyperparameter to adjust the mask ratio. The original formulation~\cite{qian2020glancing} utilized a deterministic mapping ($g$), while we use a stochastic function such as Poisson distribution to enable sampling a wider range of lengths including ``no glancing''.

The original GLAT~\cite{qian2020glancing} assumes to work with golden length so that it can glance the target by placing the target word embedding to the corresponded inputs. It is natural to use GLAT for models with CE or AXE loss, while incompatible with CTC as we always require the inputs longer than the targets.
To enable GLAT training, 
we glance target tokens from the viterbi aligned tokens $\bm{\hat{a}}=\argmax_{\bm{a}\in \Gamma(\bm{y})}p_\theta(\bm{a}|\bm{x})$ which has the same length as the decoder inputs.

Intuitively, the poorly trained model will observe more target tokens.  When the model becomes better and generate higher quality sequences, the number of masked words will be larger, which helps the model gradually learn generating the whole sentence.

\subsection{Summary}
To sum up, we conclude all proposed aspects with methods in Table~\ref{Tab:bookRWCal} which are  able to effectively help close the performance gap for fully non-autoregressive models. Note that, although none of these solutions are perfect that reduces the harmful dependency in the output space of NAT models. In practice, we find these methods target on different aspects of \textit{dependency reduction} and indicates the improvements may be complementary.

\section{Experiments} \label{experiment}
We perform extensive experiments on three challenging translation datasets by combining all mentioned techniques to check (1) whether the proposed aspects for \textit{dependency reduction} are complementary; (2) how much we can minimize the gap between a fully non-autoregressive model with the autoregressive counterpart. 

\subsection{Experimental Setup}
\label{4-1}

\paragraph{Dataset and Preprocessing}
We validate our proposed models on three standard translation benchmarks with variant sizes, i.e., WMT14 English (EN) $\leftrightarrow$ German (DE) (4.0M pairs), WMT16 English (EN) $\leftrightarrow$ Romanian (RO) (610k pairs) and WMT20 Japanese (JA) $\rightarrow$ English (EN) (13M pairs after filtering). 
For EN$\leftrightarrow$DE and EN$\leftrightarrow$RO, 
we apply the same prepossessing steps and learn sub-words as mentioned in prior work (EN$\leftrightarrow$DE:~\citealp{zhou2019understanding}, EN$\leftrightarrow$RO:~\citealp{lee2018deterministic}). 
For JA$\rightarrow$EN, the original data (16M pairs) is first filtered with Bicleaner~\cite{prompsit:2018:WMT}~\footnote{\url{https://github.com/bitextor/bicleaner}} and we apply SentencePiece~\cite{kudo-richardson-2018-sentencepiece} to generate 32,000 subwords. 

\paragraph{Knowledge Distillation} Following previous efforts, the NAT model in our experiments is trained on distilled data generated from pre-trained transformer models (\textit{base} for WMT14 EN$\leftrightarrow$DE and WMT16 EN$\leftrightarrow$RO and \textit{big} for WMT20 JA$\rightarrow$EN) using beam search with a beam size $5$.

\paragraph{Decoding}\label{3_5}
At inference time, the most straightforward way is to generate the sequence with the highest probability at each position. The outputs from the CTC-based NAT models require additional collapse process $\Gamma^{-1}$ which can be done instantly.

A relatively more accurate method is to decode multiple sequences, and rescore them to obtain the best candidate in parallel, i.e. \textit{noisy parallel decoding}~\citep[NPD,][]{gu2017non}. 
Furthermore, CTC-based models are also capable of decoding sequences using beam-search~\cite{libovicky-helcl-2018-end}, and optionally combined with $n$-gram language models~\cite{heafield-2011-kenlm,kasner2020improving}. More precisely, we search in a beam to approximately find the optimal $\bm{y}^*$ that maximizes:
\begin{equation}
    \log p_\theta(\bm{y}|\bm{x}) + \alpha \cdot \log p_{\text{LM}}(\bm{y}) + \beta \log |\bm{y}|
    \label{eq.ctc_beam}
\end{equation}
where $\alpha$ and $\beta$ are hyperparameters for language model scores and word insertion bonus.
In principle, it is no longer non-autoregressive as beam-search is a sequential process by nature. It does not contain any neural network computations and can be implemented efficiently in C++~\footnote{\url{https://github.com/parlance/ctcdecode}}. 

\begin{table*}[th!]
\centering
\small
\caption{Performance comparison between our models and existing methods. The speed-up is measured on WMT14 En-De test set. All results reported standalone are without re-scoring. \textbf{Iter.} denotes the number of iterations at inference time, \textbf{Adv.} means adaptive, $^*$ denotes models trained with distillation from a \textit{big} Transformer.}
\begin{tabular}{llcrcccc}
\toprule
 \multicolumn{2}{l}{\multirow{2}{*}{\textbf{Models}}} & 
 \multirow{2}{*}{\textbf{Iter.}} &
 \multirow{2}{*}{\textbf{Speed}}&
 \multicolumn{2}{c}{\textbf{WMT'14}} & \multicolumn{2}{c}{\textbf{WMT'16}} \\
 \multicolumn{2}{c}{} & & & \textbf{EN-DE} & \textbf{DE-EN} & \textbf{EN-RO} & \textbf{RO-EN} \\
\midrule
\multirow{3}{*}{AT}
& Transformer \textit{base} (teacher)& N & 1.0$\times$ &  \bf{27.48}&\bf{31.39}&\bf{33.70}&\bf{34.05} \\
& Transformer \textit{base} (12-1) & N & 2.4$\times$ & 26.21& 30.80& 33.17 &33.21\\
& \quad + KD & N & 2.5$\times$ & 27.34& 30.95 &33.52 &34.01\\
\cmidrule[0.6pt](lr){1-8}
\multirow{6}{*}{Iterative NAT}
& iNAT~\cite{lee2018deterministic} & 10 & 1.5$\times$ & 21.61 & 25.48 & 29.32 & 30.19 \\
& Blockwise~\cite{stern2018blockwise} & $\approx N / 5$ & 3.0$\times$  & 27.40 & - & - & -\\
& InsT~\cite{stern19a} & $\approx\log N$  & 4.8$\times$ & 27.41 & - & - & \\
& CMLM~\cite{ghazvininejad2019mask}$^*$ & 10 & 1.7$\times$ & 27.03 & 30.53 & 33.08 & 33.31 \\
& LevT~\cite{gu2019levenshtein} & Adv. & 4.0$\times$ & 27.27 & - & - & 33.26 \\
& KERMIT~\cite{chan2019kermit} &$\approx\log N$  & - & 27.80 & 30.70 & - & -\\
& LaNMT~\cite{shu2020latent} & 4 & 5.7$\times$ & 26.30 & - & - & 29.10 \\
& SMART~\cite{ghazvininejad2020semi}$^*$ & 10 & 1.7$\times$ & 27.65 & 31.27 & - & -\\
& DisCO~\cite{kasai2020non}$^*$ & Adv. & 3.5$\times$ & 
27.34 & 31.31 & 33.22 & 33.25  \\
& Imputer~\cite{saharia-etal-2020-non}$^*$ & 8& 3.9$\times$ & \bf{28.20} & \bf{31.80} & \bf{34.40} & \bf{34.10} \\
\cmidrule[0.6pt](lr){1-8}
\multirow{17}{*}{Fully NAT}
& Vanilla-NAT~\cite{gu2017non} &  1 & 15.6$\times$ & 17.69 & 21.47 & 27.29 & 29.06 \\
& LT~\cite{kaiser2018fast} & 1 & 3.4$\times$ & 19.80 & - & - & - \\
& CTC~\cite{libovicky-helcl-2018-end} & 1 & - & 16.56 & 18.64 & 19.54 & 24.67 \\
& NAT-REG~\cite{wang2019non} & 1 & - & 20.65 & 24.77 & - & -\\
& Bag-of-ngrams~\cite{shao2020minimizing} & 1 & 10.0$\times$ & 20.90 & 24.60 & 28.30 & 29.30\\
& Hint-NAT~\cite{li2018hint} & 1 & - & 21.11 & 25.24 & - & -\\
& DCRF~\cite{sun2019fast} & 1 & 10.4$\times$ & 23.44 & 27.22 &  - & -\\
& Flowseq~\cite{ma2019flowseq} & 1 & 1.1 $\times$ & 23.72 &  28.39 & 29.73  & 30.72\\
& ReorderNAT~\cite{ran2019guiding} & 1 & 16.1$\times$ & 22.79 & 27.28 & 29.30 & 29.50 \\
& AXE~\cite{GhazvininejadKZ20}$^*$ & 1 & 15.3$\times$ & 23.53 & 27.90 & 30.75 & 31.54 \\
& EM+ODD~\cite{sun2020approach} & 1 & 16.4$\times$ &24.54 & 27.93 & - & -\\
& GLAT~\cite{qian2020glancing} & 1 & 15.3$\times$  & 25.21 & 29.84 & 31.19 & 32.04 \\
& Imputer~\cite{saharia-etal-2020-non}$^*$ & 1 & 18.6$\times$  & 25.80 & 28.40 & 32.30 & 31.70 \\
\cmidrule[0.6pt](lr){2-8}
& \textbf{\textit{Ours (Fully NAT)}} & 1 & 17.6$\times$& 11.40 & 16.47 & 24.52 & 24.79\\
& \quad + KD & 1 & 17.6$\times$& 19.50 & 24.95 & 29.91 & 30.25\\
& \quad + KD + CTC & 1& 16.8$\times$& 26.51 & 30.46 & 33.41 & 34.07\\
& \quad + KD + CTC + VAE & 1 & 16.5$\times$& \bf{27.49} & 31.10 & \bf{33.79} & 33.87\\
& \quad + KD + CTC + GLAT& 1 & 16.8$\times$& 27.20 & \bf{31.39} & 33.71 & \bf{34.16}\\

\bottomrule
\end{tabular}
\label{tab:main-rst}
\end{table*}

\paragraph{Baselines}
We adopt Transformer (AT) and existing NAT approaches (see Table~\ref{tab:main-rst}) for comparison.
For Transformers, except for using the standard \textit{base} and \textit{big} architectures~\cite{vaswani2017attention} as baselines, we also compare with a \textit{deep encoder shallow encoder} Transformer suggested in \newcite{kasai2020deep} that follows the parameterizations of \textit{base} with 12 encoder layers and 1 decoder layer (i.e. \textit{base} (12-1) for short). 

\paragraph{Evaluation}
BLEU~\cite{papineni2002bleu} is used to evaluate the translation performance for all models. Following prior works, we compute tokenized BLEUs for EN$\leftrightarrow$DE and EN$\leftrightarrow$RO, while using SacreBLEU~\cite{post-2018-call} for JA$\rightarrow$EN.

In this work, we use three measures to fully investigate the translation latency of all the models:
\begin{itemize}[leftmargin=*]
    \vspace{-5pt}
    \item $\mathcal{L}_1^{\text{GPU}}$: translation latency by running the model with one sentence at a time on a single GPU. For most of existing works on NAT models, $\mathcal{L}_1^{\text{GPU}}$ is \textit{de-facto} which aligns applications like instantaneous machine translation.\vspace{-5pt}
    \item $\mathcal{L}_1^{\text{CPU}}$: the same as $\mathcal{L}_1^{\text{GPU}}$ while running the model without GPU speed-up. Compared to $\mathcal{L}_1^{\text{GPU}}$, it is less friendly to NAT models that make use of parallelism, however, closer to real scenarios.\vspace{-5pt}
    \item $\mathcal{L}_{\max}^{\text{GPU}}$: the same as $\mathcal{L}_1^{\text{GPU}}$ on GPU while running the model in a batch with as many sentences as possible. In this case, the hardware memory bandwidth are taken into account.\vspace{-1pt}
\end{itemize}
We measure the wall-clock time for translating the whole test set, and report the averaged time over sentences as the latency measure.

\begin{table*}[t]
\centering
\small
\caption{Performance comparison between fully NAT and AT models on WMT'20 JA$\rightarrow$EN. Translation latency on both GPU and CPUs are reported over the test set. The brevity penalty (BP) is also shown for reference.
}
\begin{tabular}{llccrrrr}
\toprule
& Configuration & BLEU ($\Delta$) & BP 
& \multicolumn{2}{c}{$\mathcal{L}_1^{\text{GPU}}$ (Speed-up)} 
& \multicolumn{2}{c}{$\mathcal{L}_1^{\text{CPU}}$ (Speed-up)} 
\\
\midrule
\multirow{3}{*}{AT} 
&\textit{big} \ \ (teacher) & 21.07 & 0.920 &  345 ms & 1.0 $\times$ & 923 ms & 1.0 $\times$ \\
&\textit{base} & 18.91 & 0.908 & 342 ms & 1.0 $\times$ & 653 ms & 1.4 $\times$\\
&\textit{base} (12-1) & 15.47 & 0.806 & 152 ms & 2.3 $\times$ & 226 ms & 4.0 $\times$\\
&\textit{base} (12-1) + KD & 18.76 & 0.887 & 145 ms & 2.4 $\times$ & 254 ms & 3.6 $\times$\\
\midrule
\multirow{6}{*}{NAT}
& KD + CTC & 16.93  (+0.00) & 0.828 & 17.3 ms & 19.9 $\times$ & 84 ms & 11.0 $\times$\\
& KD + CTC + VAE & 18.73 (+1.80)& 0.862 & 16.4 ms  & 21.0 $\times$ & 83 ms & 11.1 $\times$\\
&\quad w. \textit{BeamSearch20} & 19.80 (+2.87) & 0.958& 28.5 ms & 12.1 $\times$ & 99 ms & 9.3 $\times$\\
&\quad w. \textit{BeamSearch20} + \textit{4-gram LM} & \bf{21.41 (+4.48)}& 0.954& 31.5 ms & 11.0 $\times$ & 106 ms & 8.7 $\times$\\
&\quad w. \textit{NPD5} & 18.88 (+1.95)& 0.866& 34.9 ms & 9.9 $\times$ &  313 ms & 2.9 $\times$\\
&\quad w. \textit{NPD5} + \textit{BeamSearch20} + \textit{4-gram LM} & \bf{21.84 (+4.91)}& 0.962 & 57.6 ms & 6.0 $\times$ & 284 ms & 3.2 $\times$\\
\bottomrule
\end{tabular}

\label{tab:enja-rst}
\end{table*}
\begin{figure*}[t]
\centering
\subfigure{
\includegraphics[width=0.31\textwidth]{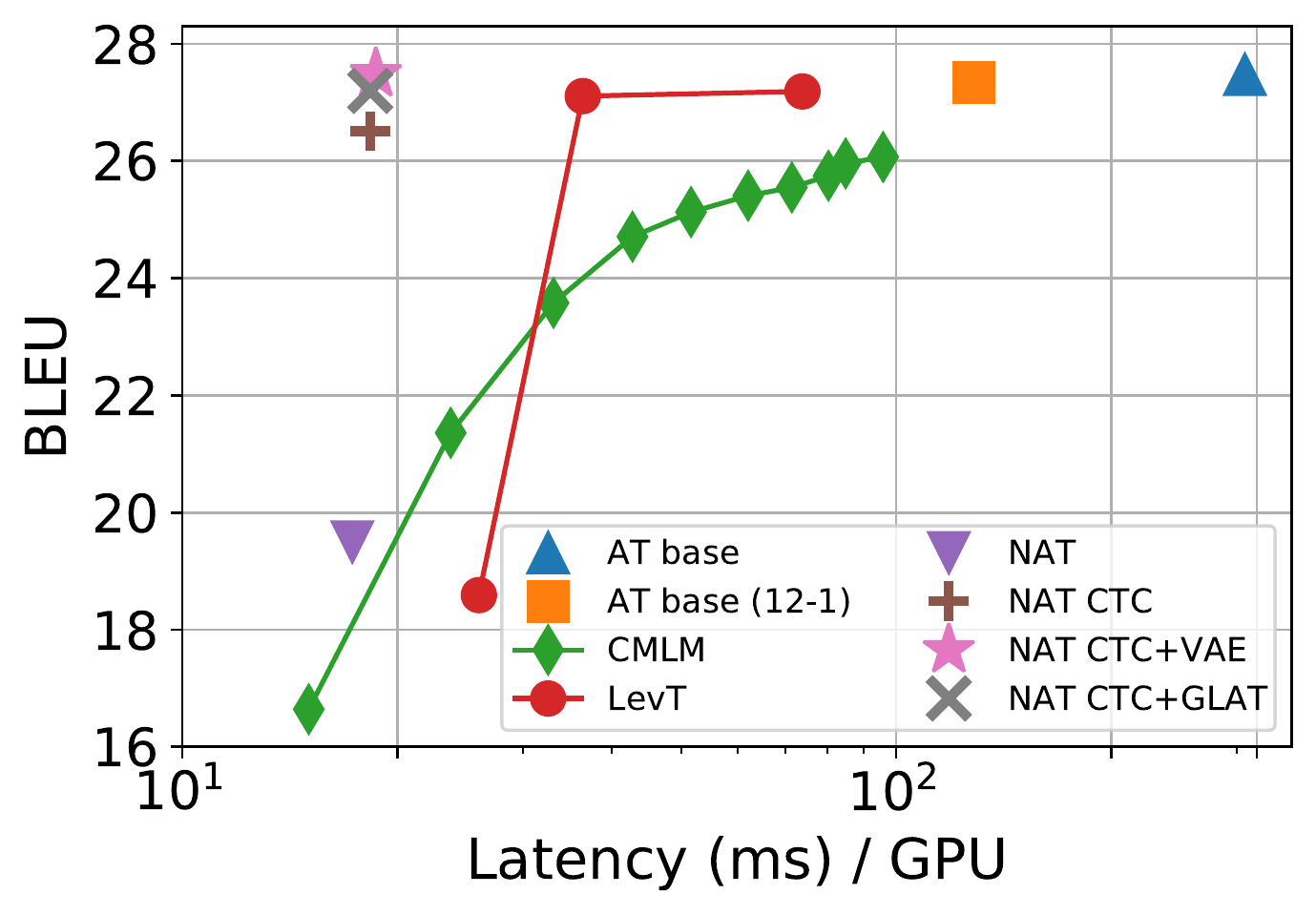}
}
\subfigure{
\includegraphics[width=0.31\textwidth]{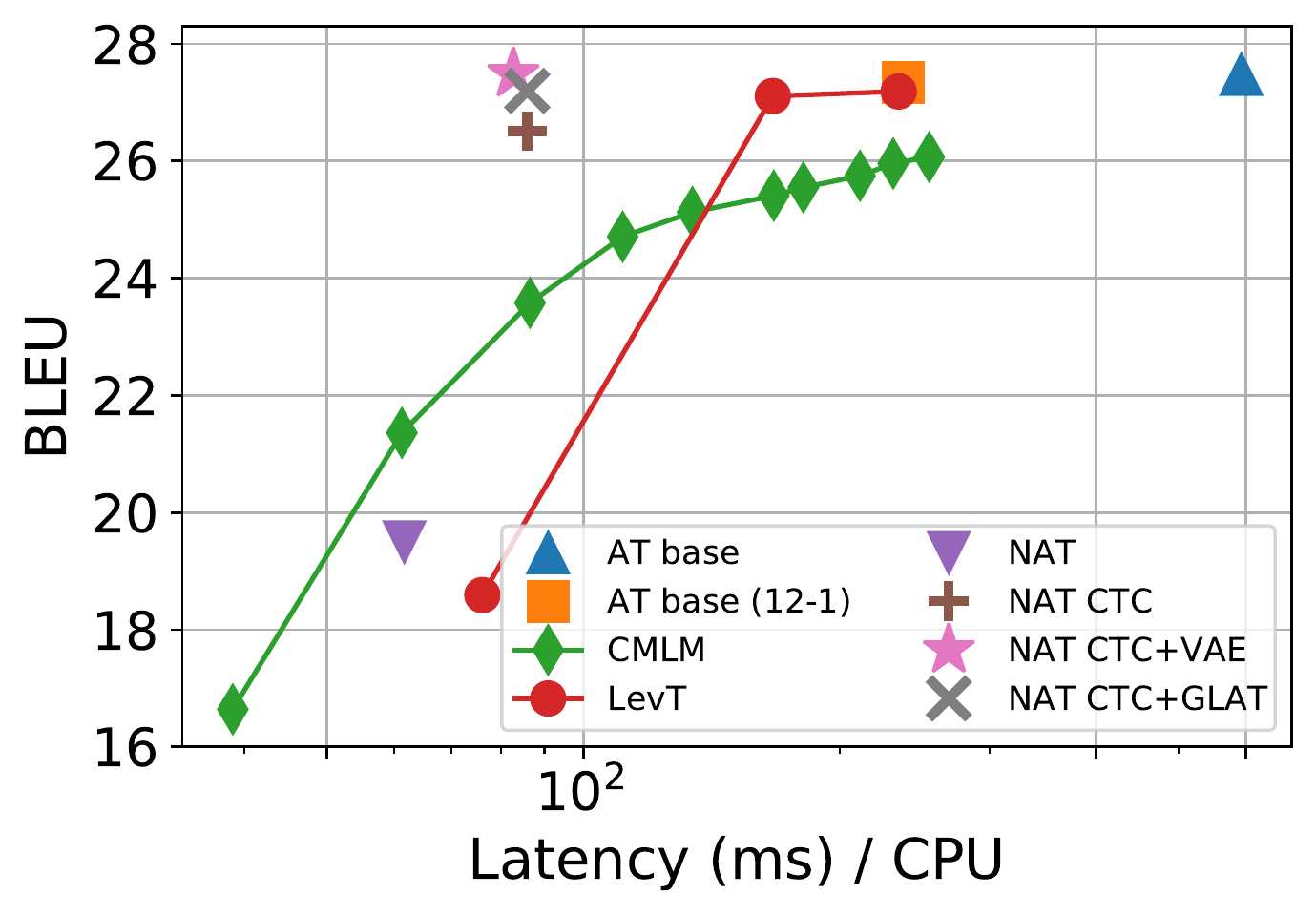}
}
\subfigure{
\includegraphics[width=0.31\textwidth]{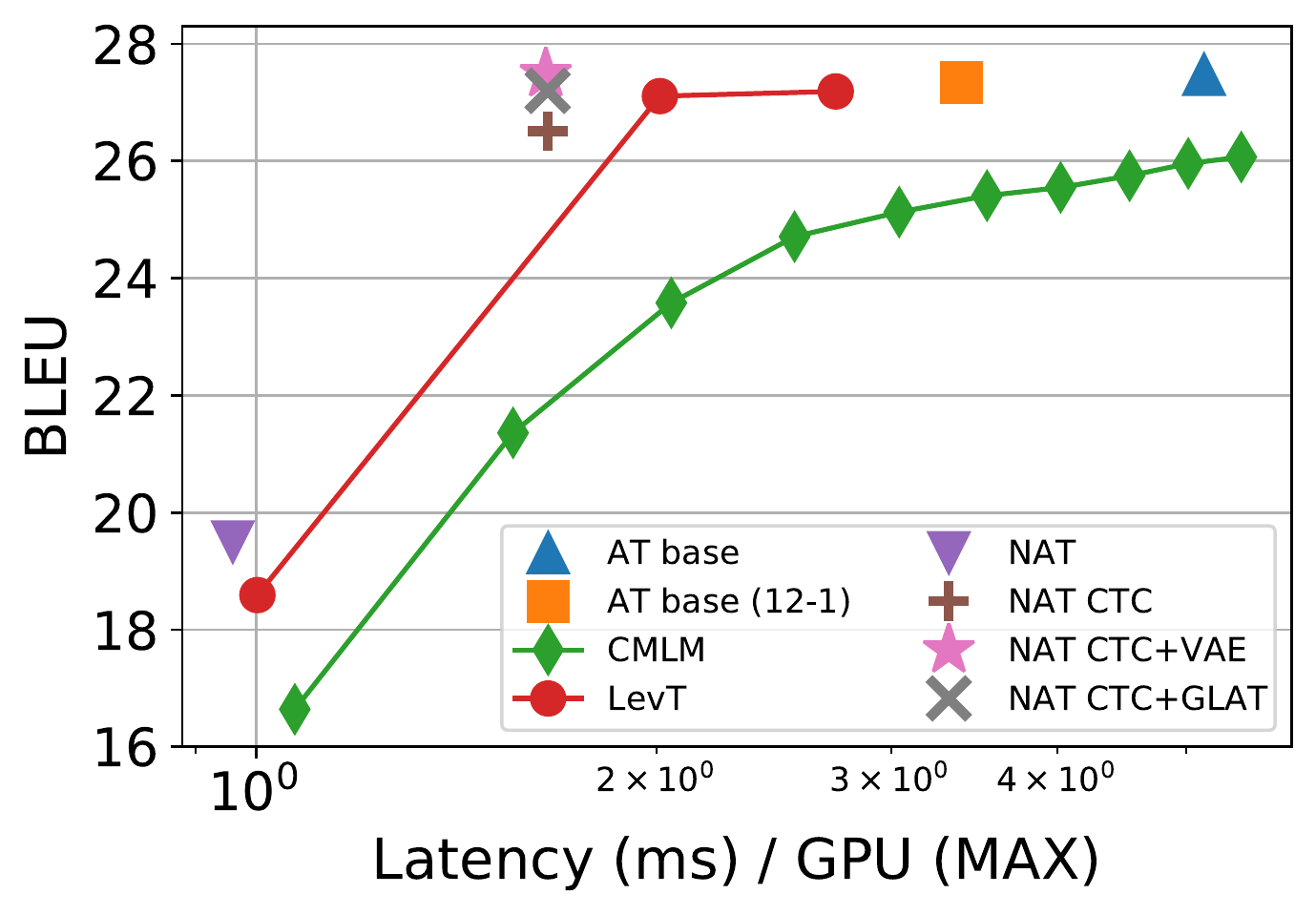}
}
\caption{Quality v.s. Latency (the upper left corner achieves the best trade-off) for fully NAT models with other translation models (AT \textit{base} and \textit{base} 12-1~\cite{kasai2020deep}, CMLM~\cite{ghazvininejad2019mask} and LevT~\cite{gu2019levenshtein}) on WMT'14 EN$\rightarrow$DE.
We evaluate latency in three setups (from left to right: $\mathcal{L}_1^\text{GPU}$, $\mathcal{L}_1^\text{CPU}$, $\mathcal{L}_{\max}^\text{GPU}$) and show them in Logarithmic scale for better visualization.
\vspace{-10pt}}
\label{fig:speed_compare}
\end{figure*}

\paragraph{Implementation Details}
We design our fully NAT model with the hyperparameters of the \textit{base} Transformer: 8-512-2048~\cite{vaswani2017attention}. 
For EN$\rightarrow$DE experiments, we also implement the NAT model in \textit{big} size: 8-1024-4096 for comparison.
For experiments using variational autoencoders (VAE), we use the last layer encoder hidden states to predict the mean and variance of the prior distribution. The latent dimension $D$ is set to $8$, and the predicted $\bm{z}$ are linearly projected and added on the encoder outputs.  Following~\newcite{shu2020latent}, we use a $3$ layer encoder-decoder as the posterior network, and apply freebits annealing~\cite{chen2016variational} to avoid posterior collapse. 
By default, we set the length for decoder inputs $3\times$ as long as the source for CTC, while using the golden length for other objectives (CE, AXE). We also learn an additional length predictor when CTC is not used. For both cases, we use \textit{SoftCopy}~\cite{wei2019imitation} which interpolated the encoder outputs as the decoder inputs based on the relative distance of source and target positions. We choose 
mask ratio $f_\text{ratio}=0.5$ for GLAT training.

For both AT and NAT models, we set the dropout rate $0.3$ for EN$\leftrightarrow$DE and EN$\leftrightarrow$RO, and $0.1$ for JA$\rightarrow$EN. We apply weight decay $0.01$ as well as label smoothing $\epsilon=0.01$. All models are trained for $300$K updates using Nvidia V100 GPUs with a batch size of approximately $128$K tokens. We measure the validation BLEU scores for every $1000$ updates, and average the best $5$ checkpoints to obtain the final model. 
We measure the GPU latency by running the model 
on a single Nvidia V100 GPU, and CPU latency on Intel(R) Xeon(R) CPU E5-2698 v4 @ 2.20GHz 
with 80 cores. 
All models are implemented on \texttt{fairseq}~\cite{ott2019fairseq}.

\subsection{Results}\label{4-2}
\paragraph{WMT'14 EN$\leftrightarrow$DE \& WMT'16 EN$\leftrightarrow$RO}
We report the performance of our fully NAT model comparing with AT and existing NAT approaches (including both iterative and fully NAT models) in Table~\ref{tab:main-rst}. Iterative NAT models with enough number of iterations generally outperform fully NAT baselines by a certain margin as they are able to recover the generation errors by explicitly modeling dependencies between (partially) generated tokens. However, such speed advantage is relatively marginal compared to AT \textit{base} (12-1) which also achieves $2.5$ times faster than the AT baseline.

Conversely,
our fully NAT models are able to readily achieve over $16$ times speed-up on EN$\rightarrow$DE by restricting translation with a single iteration. Surprisingly, simply training NAT with KD and CTC loss already beats the state-of-the-art for a single iteration NAT models across all four directions.
Moreover, combining with either latent variables (VAE) or glancing targets (GLAT) further closes the performance gap or even outperforms the AT results on both language pairs.
For example, our best model achieves \textbf{27.49} BLEU on WMT14 EN-DE -- almost identical to the AT performance (27.48) while $\bf{16.5}$ times faster in the inference time. 
Also, Table~\ref{tab:main-rst} indicates the difficulties of learning NAT on each dataset. For instance, EN$\leftrightarrow$RO is relatively easier where ``KD + CTC'' is enough to reduce target dependencies, while by contrast, applying VAE or GLAT helps to capture non-monotonic dependencies and improve by $0.5\sim1$ BLEU points on EN$\leftrightarrow$ED.
For both datasets, we ONLY need single greedy generation to achieve similar translation quality as AT beam-search results.  

\paragraph{WMT'20 JA$\rightarrow$EN} 
We also present results for training the fully NAT model on a more challenging benchmark of WMT'20 JA$\rightarrow$EN which is much larger (13M pairs) and noisier. In addition, JA is linguistically distinct to EN which makes it harder to learn mappings between them. Consequently, both AT (12-1) and our fully NAT models (see Table~\ref{tab:enja-rst}) become less confident and tend to produce shorter translations (BP $<$ 0.9), and in turn underperform the AT teacher even trained with KD. 
\paragraph{Beam search \& NPD} The performance of our NAT can be further boosted by allowing additional searching (beam-search) or re-ranking (NPD) after prediction. For CTC beam search, we use a fixed beam-size $20$ while grid-search $\alpha, \beta$ (Eq.\eqref{eq.ctc_beam}) based on the performance on the validation set. The language model~\footnote{\url{https://github.com/kpu/kenlm}} is trained directly on the distilled target sentences to avoid introducing additional information. 
For noisy parallel decoding (NPD), we draw multiple $\bm{z}$ from the learned prior distribution~\footnote{We multiply $0.1$ to the variance for less noisy samples.}, and use the teacher model (AT \textit{big}) to rerank the best $\bm{z}$ with the corresponded translation. 

As shown in Table~\ref{tab:enja-rst}, in spite of similar GPU latency ($\mathcal{L}_1^{\text{GPU}}$), beam search is much more effective than NPD with re-ranking, especially combined with a 4-gram LM where we achieve a BLEU score of $21.41$, beating the teacher model with $11\times$ speed-up. More importantly, by contributing the insertion bonus (3rd term in Eq~\eqref{eq.ctc_beam}) with $\beta$ in beam search, we have the explicit control to improve BP and output longer translations.
Also, we gain another half point by combining NPD and beam search. To have a fair comparison, we also report latency on CPUs where it is limited to leverage parallelism of the device. The speed advantage drops rapidly for NAT models, especially for NAT with NPD, however, we still maintain around 100 ms latency via beam search -- over 2$\times$ faster than the AT (12-1) systems with higher translation quality.

\paragraph{Quality v.s. Latency}
We perform a full investigation for the trade-off between translation quality and latency 
across AT, iterative NAT and our fully NAT models. The results are plotted in Figure~\ref{fig:speed_compare}. For fully NAT models, no beam search or NPD is considered.
In all three setups, our fully NAT models obtain superior trade-off compared with AT and iterative NAT models. 
Iterative NAT models (LevT and CMLM) require multiple iterations to achieve reliable performance with the sacrifice of latency, especially for $\mathcal{L}_1^\text{CPU}$ and $\mathcal{L}_{\max}^\text{GPU}$ where iterative NAT performs similarly or even worse than AT \textit{base} (12-1), leaving fully NAT models a  better position in quality-latency trade-off.  

Figure~\ref{fig:speed_compare} also shows the speed advantage of fully NAT models shrinks in the setup of $\mathcal{L}_1^\text{CPU}$ and $\mathcal{L}_\text{max}^{\text{GPU}}$ where parallelism is constrained. NAT models particularly those trained with CTC cost more computations and memory compared to AT models with a shallow decoder. For instance when calculating $\mathcal{L}_\text{max}^{\text{GPU}}$, we notice that the maximum allowed batch is 120K tokens for AT \text{base} (12-1), while we can only compute 15K tokens at a time for NAT with CTC due to the up-sampling step, even though the NAT models win the actual wall-clock time. We mark it as one limitation for future research.


\subsection{Ablation Study}
\paragraph{Impact of variant techniques}
Our fully NAT models benefit from many dependency reduction techniques in four aspects (data, model, loss function and learning). 
We analyze their effect on translation accuracy through various combinations in Table~\ref{tab:combine_tech}. First of all, 
The combination without KD has a clear performance drop compared to the one with KD, showing its high importance in NAT training. For the loss function, although AXE and CTC both consider latent alignments between target positions, the CTC-based model obtains better accuracy by marginalizing all valid alignments. Incorporating latent variables also effectively improve the accuracy moderately (around 1 BLEU). Because of the capacity to reduce the mismatch between training and inference time, the model with GLAT is more superior to the one with the RNG training method. Moreover, we find that the KD and CTC are fundamental components to build a robust NAT model. Adding either VAE or GLAT to them achieve similar performance as AT models.


\begin{table}[t]
    \small
    \centering
    \caption{Ablation on WMT14 EN$\rightarrow$DE test set with different combinations of techniques. The default setup shows a plain NAT model~\cite{gu2017non} directly trained on raw targets with cross entropy (CE) loss.}
    \begin{tabular}{c|cc|c|cc|c}
        \toprule
        KD & AXE & CTC & VAE & RND & GLAT & BLEU\\
        \midrule
        & & & & & & 11.40 \\
         \checkmark & & & & & &  19.50\\
        & \checkmark & & & & & 16.59 \\
         \checkmark&\checkmark & & & & & 21.66\\
        & &\checkmark & & & & 18.18 \\
        \checkmark& &\checkmark & & & & \bf{26.51}\\
        
        \midrule
        &&\checkmark &\checkmark & & & 23.58\\
        \checkmark&\checkmark & &\checkmark & & & 22.19\\
        \checkmark&&\checkmark  &\checkmark & & &\bf{27.49}\\

        \midrule
        \checkmark&\checkmark& & & \checkmark&& 22.74\\
        \checkmark&\checkmark& & & & \checkmark& 24.67\\
        \checkmark&&\checkmark & & \checkmark&& 26.16\\
         && \checkmark& & &\checkmark& 21.81 \\
        \checkmark&&\checkmark & & & \checkmark& \bf{27.20}\\
        \midrule
        \checkmark&&\checkmark &\checkmark & &\checkmark &\bf{27.21}\\
        
    \bottomrule
    \end{tabular}
    \label{tab:combine_tech}
\end{table}


\begin{table}[t]
\centering
\small
\caption{Performance comparison between AT and NAT models on the test set of WMT'14 EN$\rightarrow$DE. The latency is measured one sentence per batch and compared with the Transformer \textit{base}. For NAT model, we adopt CTC+VAE as the basic configuration.
}
\begin{tabular}{ll|cccc}
\toprule
\multicolumn{2}{c|}{\multirow{2}{*}{Models}} &\multicolumn{2}{c}{Distillation} & \multirow{2}{*}{BLEU} & \multirow{2}{*}{Speed-up} \\
&& \textit{base} & \textit{big}\\
\midrule
\multirow{5}{*}{AT} 
& \textit{base} & & & 27.43 & 1.0$\times$ \\
& \textit{big}  & & & 28.14 & 0.9$\times$\\
\cmidrule[0.6pt](lr){2-6}
& \multirow{1.7}{*}{\textit{base}} & & & 26.12 & 2.4$\times$\\
& \multirow{1.7}{*}{(12-1)}& \checkmark & & 27.34 & 2.5$\times$\\
& & & \checkmark & 27.83 & 2.4$\times$\\
\midrule
\multirow{4}{*}{NAT}
& \multirow{3}{*}{\textit{base}} & && 23.58 & 16.5$\times$\\
& & \checkmark & & 27.49 & 16.5$\times$\\
& & & \checkmark & 27.56 & 16.5$\times$\\
& \multirow{1}{*}{\textit{big}}
& & \checkmark & \bf{27.89} & 15.8$\times$\\
\bottomrule
\end{tabular}

\label{tab:big-distill}
\end{table}
\paragraph{Distillation corpus}
According to the ~\newcite{zhou2019understanding}, the best teacher model is relevant to the capacity of NAT models. 
We report the performance of models trained on real data and distilled data generated from AT \textit{base} and \textit{big} models in Table~\ref{tab:big-distill}. 
For \textit{base} models, both AT (12-1) and NAT achieve better accuracy with distillation, while AT benefits more by moving from 
\textit{base} to \textit{big} distilled data.
On the contrary, the NAT model improves marginally indicating that in terms of the modeling capacity, our fully NAT model is still worse than AT model even with 1 decoder layer. It is not possible to further boost the NAT performance by simply switching the target to a better distillation corpus. Nonetheless, it is possible simultaneously increase the NAT capacity by learning in \text{big} size. As shown in Table~\ref{tab:big-distill}, we can achieve superior accuracy compared to AT (12-1) with little effect on the translation latency ($\mathcal{L}_1^\text{GPU}$). 



\begin{table}[t]
    \small
    \centering
    \caption{Performance comparison of different upsample ratios ($\lambda$) for CTC-based models on WMT14 EN-DE test set. All models are trained on distilled data.}
    \begin{tabular}{c|cccc}
        \toprule
        $\lambda$ & BLEU & $\mathcal{L}_{1}^{\text{GPU}}$ & $\mathcal{L}_{\max}^{\text{GPU}}$ & $\mathcal{L}_{1}^{\text{CPU}}$\\
        \midrule
        1.5 & 26.16 &   17.9 ms & \bf{0.95} ms & \bf{66.6} ms\\
        2.0 & 26.39 &   17.5 ms & 1.03 ms & 71.6 ms\\
        2.5 & \bf{26.54} & 17.6 ms& 1.16 ms & 76.9 ms\\
        3.0 & 26.51 & \bf{17.0} ms & 1.32 ms & 81.8 ms\\
        \bottomrule
    \end{tabular}
    \vspace{-10pt}
    \label{tab:upsample_ratio}
\end{table}
\paragraph{Upsampling Ratio ($\lambda$) for CTC Loss}
To meet the length requirements in CTC loss, we upsample the encoder output by a factor of $3$ in our experiments. We also explore other possible values and report the performance in Table~\ref{tab:upsample_ratio}. The higher upsampling ratio provides a larger alignment space, leading to better accuracy. Nevertheless, with a large enough sampling ratio, further increase will not lead to the performance increase. Because of the high degree of parallelism,  $\mathcal{L}_{1}^{\text{GPU}}$ speed is similar among these ratios. However, the model with a  larger ratio has a clear latency drop on CPU or GPU with large batches. 



\section{Discussion and Future work} \label{discussion}
In this section, we go through the proposed four techniques again for fully NAT models. In spite of the success to close the gap with autoregressive models on certain benchmarks, we still see limitations when using non-autoregressive systems as mentioned in Table~\ref{Tab:bookRWCal}.

We and most of the prior research have repeatedly found that knowledge distillation (KD) is the indispensable \textit{dependency reduction} components, especially for training fully NAT models. Nevertheless, we argue that due to the model agnostic property, KD may lose key information that is useful for the model to translate. Moreover, \newcite{anonymous2021understanding} pointed out KD does cause negative effects on lexical choice errors for low-frequency words in NAT models.
Therefore, an alternative method that improves the training of NAT models over raw targets using such as GANs~\cite{binkowski2019high} or domain specific discriminators~\cite{donahue2020end} might be the future direction.

Apart from KD, we also notice that the usage of CTC loss is another key component to boost the performance of fully NAT models across all datasets. As discussed in \cref{4-2}, however, the need of up-sampling constrains the usage of our model on very long sequences or mobile devices with limited memory. In future work, it is possible to explore models to hierarchically up-sample the length with a dynamic ratio to  optimize the memory usage.

Lastly, both experiments with VAE and GLAT prove that it is helpful but not enough to train NAT models with loss based on monotonic alignments (e.g. CTC) only. To work on difficult pairs such as JA-EN, it may be a better option to adopt stronger models to capture richer dependency information, such as normalizing flows~\cite{oord2017parallel,ma2019flowseq} or non-parametric approaches~\cite{gu2018search}. 

\section{Related Work} \label{related-work}
Besides iterative and fully NAT models, some other works trying to improve the decoding speed of machine translation models from other aspects. One research lien is to mix AT and NAT models up. For example, \newcite{wang-etal-2018-semi-autoregressive} proposed a semi-autoregressive model which adopted non-autoregressive decoding locally but kept the autoregressive property in global. On the contrary, \newcite{kong-etal-2020-incorporating} and \newcite{ran-etal-2020-learning} introduced a local autoregressive NAT models which retained the non-autoregressive property in global. 

Alternatively, there are also some works trying to improve the decoding speed of AT models directly. For example, model quantization and pruning have been widely studied as a way to improve the decoding speed~\cite{see-etal-2016-compression,junczys-dowmunt-etal-2018-marian,kim-etal-2019-research,aji-heafield-2020-compressing}. Also  teacher-student training can improve the translation accuracy of the student model with faster decoding speed~\cite{junczys-dowmunt-etal-2018-marian}.
\section{Conclusion} \label{conclusion}
In this work, we aim to minimize the performance gap between fully NAT and AT models. We investigate some \textit{dependency reduction} methods from four perspective and carefully unite them with some necessary revisions. Experiments on three translation benchmarks demonstrate that our proposed models achieve state-of-the-art results of fully NAT models. 


\bibliography{acl2020}
\bibliographystyle{acl_natbib}

\end{document}